\pdfoutput=1

\documentclass[11pt]{article}

\usepackage[final]{ACL2023}
\usepackage{times}
\usepackage{latexsym}

\usepackage{amsmath}
\usepackage{amsfonts}
\usepackage{mathtools}
\usepackage{amsthm}
\usepackage{bm}
\usepackage{mathrsfs}
\usepackage{graphicx}
\usepackage{tikz}

\usepackage{algorithm} 
\usepackage{algorithmic}  
\usepackage{array}
\usepackage[flushleft]{threeparttable}
\newcolumntype{?}{!{\vrule width 2pt}}

\usepackage[T1]{fontenc}

\usepackage[utf8]{inputenc}
\usepackage{multirow}

\usepackage{booktabs}
\usepackage{microtype}

\usepackage{inconsolata}

\theoremstyle{plain}
\newtheorem{theorem}{Theorem}[section]

\theoremstyle{definition}

\theoremstyle{remark}
\newtheorem{remark}[theorem]{Remark}
\DeclareMathOperator{\Tr}{Tr}

\title{Pruning Foundation Models for High Accuracy without Retraining}

\author{Pu Zhao$^1$, Fei Sun$^2$, Xuan Shen$^1$, Pinrui Yu$^1$, Zhenglun Kong$^1$, Yanzhi Wang$^1$, Xue Lin$^1$  \\
$^1$ Northeastern University \\
$^2$  Meta \\
  \texttt{\{p.zhao, shen.xu, yu.pin, kong.zhe, yanz.wang,  xue.lin\}@northeastern.edu} \\
  \texttt{feisun@meta.com}}

\begin{document}
\maketitle
\begin{abstract}
Despite the superior performance, it is challenging to deploy foundation models or large language models (LLMs) due to their massive parameters and computations.  While pruning is a promising technique to reduce model size and accelerate the inference,  the traditional pruning techniques  can hardly be applied for LLMs as they need to finetune the model on the full dataset with multiple epochs consuming massive data and hardware resources.  To deal with this problem, post-training pruning methods are proposed to prune LLMs   in one-shot without retraining.  However, their accuracy after pruning may suffer from certain performance degradation  due to the lack of retraining with massive data.  To address this issue, in this paper, we first formulate the post-training problem for layer-wise LLM compression to simultaneously prune multiple weights in LLMs. Next, we provide an optimal solution for this problem and design our post-training pruning algorithm  for both unstructured  and semi-structured sparsity.   
Our extensive experiments  demonstrate the superior performance of the proposed methods in comparison to SOTA baselines  across various LLM families including transformer-based LLMs and Mamba-based LLMs.
\underline{\textit{Code link: https://github.com/piuzha/APT}}
\end{abstract}

\section{Introduction} 

Foundation models or large language models (LLMs) have achieved remarkable performance on a variety of tasks. However,  it is challenging to deploy LLMs in practical applications due to their massive parameters and computations. 
To facilitate LLM deployment in practice, various   model compression techniques targeting LLMs including pruning \cite{hubara2021accurate,elias2023sparsegpt} and quantization \cite{dettmers2022llm,frantar2022gptq,yao2022zeroquant,xiao2023smoothquant} have been proposed to reduce memory and computation costs.

The traditional pruning techniques, which finetune or retrain models \cite{li-etal-2020-efficient-transformer} on full datasets for many epochs (i.e., pruning-aware training), are too expensive for LLMs in terms of data and GPU resources.  
Thus, post-training pruning based on  well-pre-trained models with reduced resource requirements represents a more reasonable approach for LLMs.  Notably, SparseGPT \cite{elias2023sparsegpt} is the representative post-training pruning work  with outstanding performance. It reduces memory cost by sequentially loading transformer blocks, one at a time, instead of loading the whole model.  Moreover, it reduces the data cost by using only a small amount of calibration data, eliminating the retraining process on massive data.  Besides the optimization based SparseGPT, there are some other heuristic post-training pruning methods such as \cite{sun2023wanda,zhang2024plugandplay}, achieving  accuracy close to SparseGPT.

However, the performance of SparseGPT is still sub-optimal  as it relies on the  solution of  \textit{Single Removal Problem} (SRP) \cite{singh2020woodfisher,frantar2021m}  to address the pruning  of  multiple weights, which is essentially a \textit{Multiple Removal Problem} (MRP). 
In particular,  the SRP provides the optimal solution to prune one weight at a time and  modify all other weights to compensate the pruned single weight and minimize the loss. However, as the optimal solution is  unaware of  all previous pruned weights and requires to modify all other weights including previous pruned ones (making them unpruned again),  it is at odds with the MRP for multiple pruned weights.  Thus, the optimal solution in SRP can not be directly applied to solve MRP. To successfully incorporate the SRP solution,  a series of  approximation methods are adopted  in SparseGPT,  
at the cost of certain performance degradation,  as detailed in Sec. \ref{sec:limitation}.

Different from the SRP-based SparseGPT, we directly  formulate the MRP for layer-wise LLM pruning  to simultaneously prune multiple weights in LLMs. Next, we derive the optimal solution for the MRP  problem  with detailed  analysis for its  advantages. Based on the optimal solution, we design our post-training pruning algorithms for both unstructured  and semi-structured sparsity.  
In our comprehensive evaluation, we demonstrate our superior performance in terms of perplexity and zero-shot accuracy compared  with SOTA baselines for both transformer-based and Mamba-based  LLMs (such as our 4.278 perplexity  on wikitext2 v.s. 5.698 from SparseGPT for LLaMA2-70B under 2:4 sparsity).   

Our contributions are summarized as follows: 
\begin{itemize}
 \item We directly formulate the MRP for LLM pruning, to enable simultaneous  pruning of multiple weights, covering the SRP as a special case, as detailed in Sec.~\ref{sec: problem_comp}.
 \item We derive the optimal solution for the proposed MRP.  Based on that, we design  accurate post-training pruning algorithms for both unstructured and semi-structured sparsity.
 \item Our comprehensive experiments across various LLM families (based on transformers and Mamba), model sizes, and datasets   demonstrate our superior performance compared with  the optimization-based SparseGPT and other heuristic SOTA baselines. 
\end{itemize}

\section{Background}

\subsection{Post-Training Pruning}

The post-training pruning problem   can be typically  formulated as the following, 
\begin{align} \label{eq:all_pruning_problem}
\min_{\delta \bm w} \ \ &L' (\bm w + \delta \bm w)   - L' (\bm w ) , \nonumber  \\ 
s.t. \ \ 
&  ( \bm w + \delta \bm w ) \odot \bm M = \bm 0, 
\end{align}
where $\bm w$ is the original model weights, $\delta \bm w$ is the modifications of model weights and $\bm M$ is the binary pruning mask on model weights with 1 denotes pruning. $L' (\bm w )$  is the typical training loss. The problem minimizes the difference of the loss before and after pruning, by optimizing the unpruned weights, with the constraint  that the pruned weights following the mask should be zero.

\subsection{Single Removal Problem}

The single removal problem (SRP)  is investigated in many  works \cite{singh2020woodfisher,frantar2022optimal,elias2023sparsegpt} due to its simplicity.
It optimizes the following  \cite{singh2020woodfisher},
\begin{align} \label{eq:SRP}
\min_{\delta \bm w} \ \ & \mathcal{L}(\delta \bm w)  =  \frac{1}{2} \delta \bm w \bm H \delta \bm w^T, \nonumber  \\ 
s.t. \ \ 
&  ( \delta \bm w  + \bm w )   \cdot \bm e_{t}  = 0, 
\end{align}
where  $\bm e_{t}$ is a one-hot vector denoting the pruned location for the single weight, and $\bm H$ is the Hessian matrix of weights.   $\bm w$ and $\delta \bm  w$ are formulated as  one-dimensional vectors for simplicity.  The problem prunes one weight  at a time, without any information for other previous pruned weights. 

Note that here $L' (\bm w + \delta \bm w)   - L' (\bm w )  \approx \nabla_{\bm w} L' \delta \bm w^T + \frac{1}{2} \delta \bm w \bm H \delta \bm w^T  \approx \frac{1}{2} \delta \bm w \bm H \delta \bm w^T$.  
It ignores the first-order Jacobian term $\nabla_{\bm w} L' \delta \bm w^T$ and only minimizes the second-order Hessian term $\frac{1}{2} \delta \bm w \bm H \delta \bm w^T$, by assuming that the model is well-trained and thus pruned at local minimum.

\subsection{Limitations of SRP}  \label{sec:limitation}

\subsubsection{ Zero Jacobian Assumption}
Following  SRP,  SparseGPT \cite{elias2023sparsegpt} applies the SRP solution for each linear layer in LLMs with   $\bm H = 2\bm x  \bm x^T$.  However, the assumption with  zero Jacobian  does not hold  in this layer-wise  pruning setting with a local quadratic loss $ L'(\bm w ) = \| \bm w \bm x \|_2^2 $, since we can directly obtain the Jacobian  $\nabla_{\bm w} \hat L = 2 \bm w \bm x \bm x^T$  which is non-zero. The assumption may be unreasonable here. 

\subsubsection{Sequential Weight Freezing } \label{sec:limitation2}

Another difficulty with the SRP solution is its unawareness of  all other pruned weights  during pruning.   Specifically, when compensating the loss of a single pruned weight, its optimal solution requires to modify all other weights including all previous pruned weights,  making them unpruned again and violating the pre-defined sparsity requirement. 

Thus, the SRP solution is not able to directly prune multiple weights.  To address this problem, SparseGPT applies a series of techniques such as Optimal Partial Updates and Hessian Synchronization.   
The key idea is to  sequentially prune weights in the same row, and  freeze/fix all weights (including pruned and unpruned weights) previous to the current pruned weight, so that the previous pruned weights are kept zero. The drawback is that all previous unpruned weights  are also frozen without further updating,  leading to  sub-optimal achievements with potential performance degradation. 

\section{Multiple Removal Problem}

\subsection{Notations} \label{sec_notations}

For   linear layers, the forward computation can be represented as $\bm w \bm x$,  where $\bm w \in \mathbb{R}^{n\times m}$ is the   weights and  $\bm x \in \mathbb{R}^{m\times B}$  ($B$ is the token number in each batch) is the layer input.   $[\bm A]_{q,p}$ denotes the weight in the $q^{th}$ row and the $p^{th}$ column of the 2D matrix $\bm A$.   $[\bm A]_{q,:}$ denotes the $q^{th}$ row  of $\bm A$, and   $[\bm A]_{:,p}$ represents the    $p^{th}$ column of   $\bm A$.  

To make  the problem tractable,  the pruned weights are distributed in $k$  rows ($ k\le n$), and their row indexes are denoted by  $q_i, \forall i \in \{1,\dotsc,k\}  $. 
In the  $q_i^{th}$ row, there are $k_i$ pruned elements, and their column indexes are denoted by $p_{ij}, \forall j \in \{1,\dotsc,k_i\}$. 
Since different rows $q_i$ have different  numbers and distributions  of pruned locations, we use the representation $p_{ij}$ rather than $p_j$.  
 Thus, the pruned  locations/indexes in the weight matrix $\bm w$ can be expressed as  $(q_i, p_{ij}),  \forall i  \in \{1,\dotsc,k\},  \forall j \in \{1,\dotsc,k_i\}  $, or  $(q_i, p_{i1}), (q_i, p_{i2}), ..., (q_i, p_{ik_i}),  \forall i  \in \{1,\dotsc,k\} $.  

  $\bm e_s$ is a  one-hot vector with the $s^{th}$ element as one and all others as zero. So  $\bm e_{q_i}^T  \bm w \bm e_{p_{ij}}$  means the weight in the  $q_i^{th}$ row and the $p_{ij}^{th}$ column of $ \bm w$, with $\bm e_{q_i} \in \mathbb{R}^{n\times 1}$ and $\bm e_{p_{ij}} \in \mathbb{R}^{m\times 1}$.

$\Tr(\cdot)$ represents the trace function. 

\subsection{Motivation with MRP} 

To address the limitations of the SRP,  we  try to formulate and solve the MRP, which prunes multiple weights simultaneously.  Our MRP is specifically formulated for the layer-wise LLM pruning without any assumptions. Furthermore,  since our MRP prunes multiple weights at the same time,  each pruned weight is aware of  all other pruned weights and thus there is no  need to freeze any  weights, which effectively addresses the limitations of  SRP.

\subsection{MRP Formulation}

In LLMs, the   linear layers in transformer \cite{attention} or Mamba \cite{gu2023mamba} blocks  are the main cost of computations and  parameters.  To reduce the  overhead of pruning LLMs, following SparseGPT, we adopt the layer-wise compression strategy to sequentially  load and prune one single block  instead of the whole model.   The significantly reduced memory  cost makes it feasible to use only one single GPU for all computations. 
  
For each linear layer, we try to minimize the difference of  the linear outputs  (measured by $\ell_2$ norm)  before and after pruning,  i.e.,  $\| (\bm w + \delta \bm w) \bm x - \bm w \bm x  \|_2^2 =  \| \delta \bm w  \bm x \|_2^2 $. To make  the problem tractable,  as discussed in Sec.~\ref{sec_notations}, the pruned  locations/indexes in the weight matrix $\bm w$ can be expressed as $(q_i, p_{i1}), (q_i, p_{i2}), ..., (q_i, p_{ik_i}),  \forall i  \in \{1,\dotsc,k\} $.  The pruned weights $(q_i, p_{ij})$  are set to zero, i.e.  $[\bm w + \delta \bm w]_{q_i,p_{ij}} = 0$. To minimize the loss incurred by pruning, the other unpruned weights are  updated for the compensation of pruned weights. 
Our  MRP  is  formulated as the following,

{\footnotesize
\begin{align} \label{eq_ori_problem}
\min_{\delta \bm w} \ \ & \mathcal{L}(\delta \bm w)  =   \| \delta \bm w   \bm x \|^2 , \nonumber  \\ 
s.t. \ \ 
& \bm e_{q_i}^T \delta \bm w \bm e_{p_{i1}}  + [w]_{q_i,p_{i1}} = 0,  \nonumber  \\
& \bm e_{q_i}^T \delta \bm w  \bm e_{p_{i2}} + [w]_{q_i,p_{i2}} = 0,  \nonumber  \\ 
& ...... \nonumber  \\
& \bm e_{q_i}^T \delta \bm w \bm e_{p_{ik_i}}  + [w]_{q_i,p_{ik_i}} = 0,  \nonumber  \\ 
& \forall i \in \{1,\dotsc,k\}, 
\end{align}}%
where $\bm e_{q_i}^T \delta \bm w \bm e_{p_{ij}}$  denotes the weight in the  $q_i^{th}$ row and the $p_{ij}^{th}$ column of $\delta \bm w$. 

It can be  transformed to  vector representation,
{\footnotesize
\begin{align} \label{eq_problem}
\min_{\delta \bm w} \ \ & \mathcal{L}(\delta \bm w)  =   \| \delta \bm w   \bm x \|^2 ,      \nonumber  \\
s.t. \ \  & \bm e_{q_1}^T \delta \bm w \bm e_{p\_q_{1}} + \bm w_{q_1} = \bm 0 ,  \nonumber  \\
& \bm e_{q_2}^T \delta \bm w \bm e_{p\_q_{2}} + \bm w_{q_2} = \bm 0 , \nonumber  \\
& ......   \nonumber  \\
& \bm e_{q_k}^T \delta \bm w \bm e_{p\_q_{k}} + \bm w_{q_k} = \bm 0,
\end{align}}%
where  $\bm e_{p\_{q_i}} \in \mathbb{R}^{m\times k_i}$ with $ [\bm e_{p\_{q_i}}]_{:, j} = \bm e_{p_{ij}}$, and $\bm w_{qi} = [[\bm w]_{q_i,p_{i1}}, [\bm w]_{q_i,p_{i2}}, ..., [\bm w]_{q_i,p_{ik_i}}] \in \mathbb{R}^{1\times k_i} $.  In   MRP,  multiple weights  are  pruned simultaneously.   $\bm e_{p\_{q_i}}$ is a collection of all pruned column indexes in the $q_i^{th}$ row,  and $\bm w_{qi}$  is a collection of all pruned weight values in the $q_i^{th}$ row.

\subsection{Comparison with SRP}  \label{sec: problem_comp}

Our problem formulation is different from the SRP \cite{singh2020woodfisher} in several ways. 

\textbf{Relax the zero Jacobian assumption.} 
Different from   SRP with the  zero Jacobian assumption which does not hold for the layer-wise LLM pruning,   our  formulation directly optimize the  difference  of outputs before and after pruning,   without  any assumptions or approximations.    

Furthermore, we provide an explanation for why SRP can still achieve good performance with the unreasonable  zero Jacobian assumption. Specifically, since $\bm H = 2\bm x  \bm x^T$  for  linear layers with the quadratic loss $  L'(\bm w ) = \| \bm w \bm x \|_2^2 $,  we have $\frac{1}{2} \delta \bm w \bm H \delta \bm w^T = \delta \bm w \bm x \bm x^T \delta \bm w^T = \|  \delta \bm w   \bm x \|_2^2$, which means that the optimization objective of  SRP is well aligned with that of  our proposed MRP. That is why SRP can still perform well with an unreasonable assumption. We demonstrate that our  MRP-based method can achieve better performance than the SRP  solutions such as SparseGPT.

\textbf{Simultaneous multiple removal without any approximations. }  The pruning removes multiple weights in the model. Compared with SRP, our proposed MRP directly addresses the problem by simultaneously pruning multiple weights, without the need for sequential weight freezing following a series of approximation techniques such as Optimal Partial Updates and Hessian Synchronization (see Sec.~\ref{sec:limitation2}). Our straightforward formulation leads to a direct solution to update all unpruned weights, leading to a better accuracy performance than SRP solutions which freeze part of unpruned weights without further updating. 


\textbf{Cover SRP as a special case. }
The SRP is a special case of our MRP.  Our   formulation deals with multiple weight removals in 2D weight matrices,  which covers the single weight removal within 1D weight vectors  in SRP as a special case.  Consequently,  the SRP solution is also a special case of our MRP solution.

\section{Methodology}

We first derive our optimal solution for the MRP  and then discuss the algorithm design. 

\subsection{Optimal Solution}

The Lagrange function of  Problem~(\ref{eq_problem}) is   
{\footnotesize
\begin{align} \label{equ_lagrange}
\mathscr{L}(\delta \bm w, \bm \lambda)    =  & \| \delta \bm w   \bm x \|^2   + ( \bm e_{q_1}^T \delta \bm w \bm e_{p\_q_{1}} + \bm w_{q_1} )  \bm  \lambda_{1} \nonumber \\
&+ ( \bm e_{q_2}^T \delta \bm w \bm e_{p\_q_{2}} + \bm w_{q_2} )  \bm \lambda_{2}  + ... \nonumber \\
&+  ( \bm e_{q_k}^T \delta \bm w \bm e_{p\_q_{k}} + \bm w_{q_k}  )  \bm \lambda_{k}, \nonumber  \\
  =  & \Tr ( \bm x^T \delta \bm w^T    \delta \bm w   \bm x )  +  \sum_i ( \bm e_{q_i}^T \delta \bm w \bm e_{p\_q_{i}} + \bm w_{q_i} ) \bm \lambda_i,
\end{align}}%
where  $\bm \lambda_i \in R^{k_i \times 1} $ denotes  the Lagrange multiplier corresponding to the constraint for the $q_i^{th}$ row in Problem (\ref{eq_problem}).  $\bm \lambda_i  = [ \lambda_{i1},  \lambda_{i2}, ......,  \lambda_{ik_i} ]$  and each $\lambda_{ij}$ corresponds to the constraint  $  \bm e_{q_i}^T \delta \bm w  \bm e_{p_{ij}} + [w]_{q_i,p_{ij}} = 0$ in Problem~(\ref{eq_ori_problem}). 
Unlike the SRP with a scalar  $\delta \bm w   \bm x $,  in our  problem, $ \delta \bm w   \bm x $ is a matrix, requiring the trace function $\Tr(\cdot)$.

The gradients with reference to $\delta \bm w$ should be 0.
{\footnotesize
\begin{align} 
\frac{\delta \mathscr{L}(\delta \bm w,  \bm \lambda) }{\delta  (\delta \bm w)} = 2 \delta \bm w \bm x \bm x^T + \sum_i  \bm e_{q_i} \bm \lambda_i^T \bm e_{p\_q_{i}}^T  = 0.
\end{align}}%
We can obtain $\delta \bm w$ as below, 
{\footnotesize
\begin{align} \label{equ_delta_half}
\delta \bm w   = -    \left(  \sum_i  \bm e_{q_i} \bm \lambda_i^T \bm e_{p\_q_{i}}^T  \right) (2 \bm x \bm x^T )^{-1}.
\end{align}}%

By applying Equation (\ref{equ_delta_half})  in Equation (\ref{equ_lagrange}),  we have the following, 

{\footnotesize
\begin{align}
g(\bm \lambda) =  - \frac{1}{2} \sum_i  \bm \lambda_i^T \bm e_{p\_q_{i}}^T   (2 \bm x \bm x^T )^{-1} \bm e_{p\_q_{i}} \bm \lambda_i  
+  \sum_i \bm w_{q_i} \bm \lambda_i.
\end{align}} %
Note that $\bm e_{q_i}^T \bm e_{q_i} =1 $ and $\bm e_{q_i}^T \bm e_{q_s} = 0$, for $i \neq s$.   Besides,  we can switch the  position of $  \bm x^T  (2 \bm x \bm x^T )^{-1}  \bm e_{p\_q_{i}}  \bm \lambda_i$  and $\bm \lambda_i^T \bm e_{p\_q_{i}}^T  (2 \bm x \bm x^T )^{-1} \bm x  $  in the trace function.

The gradients with reference to $\bm \lambda$ should be 0.

{\footnotesize 
\begin{align}
\frac{\delta g(\bm \lambda)}{\delta \bm \lambda_i} = -  \bm e_{p\_q_{i}}^T   (2 \bm x \bm x^T )^{-1} \bm e_{p\_q_{i}} \bm \lambda_i  + \bm w_{q_i}^T = \bm 0, \forall i.
\end{align}}%
We can obtain the optimal $\bm \lambda$ as below,
{\footnotesize
\begin{align}
\bm \lambda_i^* =    [\bm e_{p\_q_{i}}^T   (2 \bm x \bm x^T )^{-1} \bm e_{p\_q_{i}}  ]^{-1} \bm w_{q_i}^T, \forall i.
\end{align}}%

The optimal $\delta \bm w$ can be derived as below,
{\small
\begin{align}  \label{eq:optimal_w}
\delta \bm w^*  = -  & \left(  \sum_i  \bm e_{q_i}  \bm w_{q_i} [\bm e_{p\_q_{i}}^T   (2 \bm x \bm x^T )^{-1} \bm e_{p\_q_{i}}  ]^{-1}    \bm e_{p\_q_{i}}^T  \right)  \nonumber \\
& \times (2 \bm x \bm x^T )^{-1} .
\end{align} }%
The minimal loss/error corresponding to the optimal $\delta \bm w$  can be obtained by
{\footnotesize
\begin{align}  \label{eq:optimal_loss}
L^* = & \frac{1}{2}  \sum_i  \bm \lambda_i^T \bm e_{p\_q_{i}}^T   (2 \bm x \bm x^T )^{-1} \bm e_{p\_q_{i}} \bm \lambda_i  \nonumber \\
= &  \frac{1}{2}  \sum_i   \bm w_{q_i} [\bm e_{p\_q_{i}}^T   (2 \bm x \bm x^T )^{-1} \bm e_{p\_q_{i}}  ]^{-1}  \bm w_{q_i}^T.
\end{align} }

\begin{remark}
\textbf{Dampening for the inverse. } If $2 \bm x \bm x^T $ is not full rank with difficulties for the inversion $(2 \bm x \bm x^T )^{-1}$, the dampening technique is adopted  to compute $(2 \bm x \bm x^T + \gamma \bm I )^{-1} $ instead, with $\gamma$ as the dampening ratio. 
\end{remark}

\begin{remark}
\textbf{Separate row computation. } For the optimal perturbation in Equation~(\ref{eq:optimal_w}),  since $\bm e_{q_i}  $ is a one-hot vector, $\bm e_{q_i}  \times A$  only has non-zero values in the $q_i^{th}$ row with all zeros for all other rows.  Thus, in Equation~(\ref{eq:optimal_w}), each term with the index $i$ in the sum just computes the $q_i^{th}$ row in the outputs and the computation of the $q_i^{th}$ row does not affect the $q_s^{th}$ row, $ \forall s \ne i$. Specifically, we have the following, 
{\small
\begin{align} 
 [\delta \bm w^*]_{q_i,:} = -   \bm w_{q_i} [\bm e_{p\_q_{i}}^T   (2 \bm x \bm x^T )^{-1} \bm e_{p\_q_{i}}  ]^{-1}    \bm e_{p\_q_{i}}^T   (2 \bm x \bm x^T )^{-1} 
\end{align} }%
\end{remark}

\begin{remark}
\textbf{Full interactions between pruned weights. }  For our optimal perturbation in Equation~(\ref{eq:optimal_w}) and optimal loss in Equation~(\ref{eq:optimal_loss}),  our solution  is not the simple sum of  multiple SRP solutions.  Our solution not only  depends on the multiple pruned weights,  but also  takes  the interactions of the multiple removals  (denoted by  $[\bm e_{p\_q_{i}}^T   (2 \bm x \bm x^T )^{-1} \bm e_{p\_q_{i}}  ]^{-1}$) into considerations, which are unavailable in SRP without the information of other multiple removals.   
\end{remark}

\subsection{Algorithm Design} \label{sec:algorithm}

\begin{algorithm}[tb]  
   \caption{Accurate post-training pruning.}
   \label{alg:b}
\begin{algorithmic}
   \STATE {\bfseries Input:} weight matrix $\bm w$, pruning rate $\alpha$,  block-size $S$, block number $N$, $2\bm x \bm x^T + \gamma \bm I$.
   \REPEAT 
   \STATE  Initialize the pruning mask $\bm M = \bm 0$;
   \STATE  Compute the inversion $[2\bm x \bm x^T + \gamma \bm I]^{-1}$ ;
   \FOR{$i=1$ {\bfseries to} $N$}
   \STATE Find the pruned weight indexes  using either Solution $\mathfrak{S}$ or $\mathfrak{M}$ for \textit{pruning mask}; 
   \STATE Update $\bm M$ with new pruned locations;
   \STATE Compute the modifications on unpruned weights using Solution $\mathfrak{S}$ or $\mathfrak{M}$ for \textit{optimal compensation} with $\bm M$;  
   \STATE Update  unpruned  and pruned weights;
   \ENDFOR
   \UNTIL{The final linear layer is pruned.}
\end{algorithmic}
\end{algorithm}

Our   pruning algorithm is shown in Algorithm~\ref{alg:b}.  We need to address two key problems: the pruning mask and optimal compensation. For each problem, we have two choices, including  Solution $\mathfrak{M}$ from our MRP  and its simplified version, Solution $\mathfrak{S}$.

\begin{table*}[t]
\center
\scalebox{0.88}{
\begin{tabular}{ccccccccc}
\hline
\multicolumn{1}{c|}{\multirow{2}{*}{\begin{tabular}[c]{@{}c@{}}Model \&\\ Setting\end{tabular}}}  & \multicolumn{1}{c|}{\multirow{2}{*}{Datasets}} & \multicolumn{1}{c?}{\multirow{2}{*}{Origin}} & \multicolumn{2}{c?}{Unstructured 50\%}                                                                                                                   & \multicolumn{4}{c}{2:4 sparsity (50\% sparsity)}                                                                                                                                                                                                                                                          \\ \cline{4-9} 
\multicolumn{1}{c|}{}                                                                             & \multicolumn{1}{c|}{}                          & \multicolumn{1}{c?}{}                        & \multicolumn{1}{c|}{\begin{tabular}[c]{@{}c@{}}$\mathfrak{S}\mathfrak{S}$\\ (SparseGPT)\end{tabular}} & \multicolumn{1}{c?}{\begin{tabular}[c]{@{}c@{}} $\mathfrak{S}\mathfrak{M}$\\ (ours)\end{tabular}} & \multicolumn{1}{c|}{\begin{tabular}[c]{@{}c@{}}$\mathfrak{S}\mathfrak{S}$\\ (SparseGPT)\end{tabular}} & \multicolumn{1}{c|}{\begin{tabular}[c]{@{}c@{}}$\mathfrak{S}\mathfrak{M}$\\ (ours)\end{tabular}} & \multicolumn{1}{c|}{\begin{tabular}[c]{@{}c@{}}$\mathfrak{M}\mathfrak{S}$\\ (ours)\end{tabular}} & \begin{tabular}[c]{@{}c@{}}$\mathfrak{M}\mathfrak{M}$\\ (ours)\end{tabular} \\  \hline  \hline
\multicolumn{1}{c|}{\multirow{2}{*}{\begin{tabular}[c]{@{}c@{}}LLaMA2-7B\\ S=2048\end{tabular}}}  & \multicolumn{1}{c|}{wikitext2}                 & \multicolumn{1}{c?}{5.472}                   & \multicolumn{1}{c|}{7.052}                                                    & \multicolumn{1}{c?}{\textbf{7.018}}                                               & \multicolumn{1}{c|}{10.85}                                                    & \multicolumn{1}{c|}{10.15}                                               & \multicolumn{1}{c|}{10.7}                                                & \textbf{10.14}                                              \\
\multicolumn{1}{c|}{}                                                                             & \multicolumn{1}{c|}{c4}                        & \multicolumn{1}{c?}{7.263}                   & \multicolumn{1}{c|}{9.305}                                                    & \multicolumn{1}{c?}{\textbf{9.204}  }                                             & \multicolumn{1}{c|}{13.65}                                                    & \multicolumn{1}{c|}{12.48}                                               & \multicolumn{1}{c|}{13.38}                                               &\textbf{12.47}                                             \\ \hline
\multicolumn{1}{c|}{\multirow{2}{*}{\begin{tabular}[c]{@{}c@{}}LLaMA2-7B\\ S=all\end{tabular}}}   & \multicolumn{1}{c|}{wikitext2}                 & \multicolumn{1}{c?}{5.472}                   & \multicolumn{1}{c|}{7.045}                                                    & \multicolumn{1}{c?}{\textbf{7.019} }                                              & \multicolumn{1}{c|}{10.92}                                                    & \multicolumn{1}{c|}{\textbf{10.37}}                                               & \multicolumn{1}{c|}{10.6}                                                & 10.38                                               \\
\multicolumn{1}{c|}{}                                                                             & \multicolumn{1}{c|}{c4}                        & \multicolumn{1}{c?}{7.263}                   & \multicolumn{1}{c|}{9.36}                                                     & \multicolumn{1}{c?}{\textbf{9.247} }                                             & \multicolumn{1}{c|}{13.62}                                                    & \multicolumn{1}{c|}{12.762}                                              & \multicolumn{1}{c|}{13.31}                                               & \textbf{12.759}                                             \\ \hline 
\multicolumn{1}{c|}{\multirow{2}{*}{\begin{tabular}[c]{@{}c@{}}LLaMA2-13B\\ S=2048\end{tabular}}} & \multicolumn{1}{c|}{wikitext2}                 & \multicolumn{1}{c?}{4.884}                   & \multicolumn{1}{c|}{6.028}                                                    & \multicolumn{1}{c?}{\textbf{6.001} }                                              & \multicolumn{1}{c|}{8.76}                                                     & \multicolumn{1}{c|}{\textbf{8.219} }                                              & \multicolumn{1}{c|}{8.644}                                               & 8.224                                               \\
\multicolumn{1}{c|}{}                                                                             & \multicolumn{1}{c|}{c4}                        & \multicolumn{1}{c?}{6.727}                   & \multicolumn{1}{c|}{8.275}                                                    & \multicolumn{1}{c?}{\textbf{8.21}  }                                              & \multicolumn{1}{c|}{11.4}                                                     & \multicolumn{1}{c|}{10.71}                                               & \multicolumn{1}{c|}{11.28}                                               & \textbf{10.699   }                                           \\ \hline
\multicolumn{1}{c|}{\multirow{2}{*}{\begin{tabular}[c]{@{}c@{}}LLaMA2-13B\\ S=all\end{tabular}}}  & \multicolumn{1}{c|}{wikitext2}                 & \multicolumn{1}{c?}{4.884}                   & \multicolumn{1}{c|}{6.082}                                                    & \multicolumn{1}{c?}{\textbf{6.03} }                                               & \multicolumn{1}{c|}{8.732}                                                    & \multicolumn{1}{c|}{8.239}                                               & \multicolumn{1}{c|}{8.65}                                                & \textbf{8.225    }                                           \\
\multicolumn{1}{c|}{}                                                                             & \multicolumn{1}{c|}{c4}                        & \multicolumn{1}{c?}{6.727}                   & \multicolumn{1}{c|}{8.374}                                                    & \multicolumn{1}{c?}{\textbf{8.269}  }                                             & \multicolumn{1}{c|}{11.36}                                                    & \multicolumn{1}{c|}{10.796}                                              & \multicolumn{1}{c|}{11.23}                                               & \textbf{10.789 }                                             \\ \hline 
\multicolumn{1}{c|}{\multirow{2}{*}{\begin{tabular}[c]{@{}c@{}}LLaMA2-70B\\ S=all\end{tabular}}}  & \multicolumn{1}{c|}{wikitext2}                 & \multicolumn{1}{c?}{3.319}                   & \multicolumn{1}{c|}{4.509}                                                         & \multicolumn{1}{c?}{\textbf{4.142}  }                                                  & \multicolumn{1}{c|}{5.698}                                                    & \multicolumn{1}{c|}{4.278}                                               & \multicolumn{1}{c|}{4.353}                                               &\textbf{ 4.278     }                                          \\
\multicolumn{1}{c|}{}                                                                             & \multicolumn{1}{c|}{c4}                        & \multicolumn{1}{c?}{5.709}                   & \multicolumn{1}{c|}{6.932}                                                         & \multicolumn{1}{c?}{  \textbf{6.528}  }                                                  & \multicolumn{1}{c|}{8.154}                                                    & \multicolumn{1}{c|}{6.683}                                               & \multicolumn{1}{c|}{6.84}                                                & \textbf{6.683  }                                             \\ \hline    \hline
\multicolumn{1}{c|}{\multirow{2}{*}{\begin{tabular}[c]{@{}c@{}}OPT-2.7B\\ S=512\end{tabular}}}   & \multicolumn{1}{c|}{wikitext2}                 & \multicolumn{1}{c?}{12.47}                   & \multicolumn{1}{c|}{13.43}                                                    & \multicolumn{1}{c?}{\textbf{13.29}  }                                             & \multicolumn{1}{c|}{17.13}                                                    & \multicolumn{1}{c|}{16.74}                                               & \multicolumn{1}{c|}{16.89}                                               & \textbf{16.68}                                             \\
\multicolumn{1}{c|}{}                                                                            & \multicolumn{1}{c|}{c4}                        & \multicolumn{1}{c?}{14.34}                   & \multicolumn{1}{c|}{15.8}                                                     & \multicolumn{1}{c?}{\textbf{15.66} }                                              & \multicolumn{1}{c|}{19.34}                                                    & \multicolumn{1}{c|}{18.7}                                                & \multicolumn{1}{c|}{19.07}                                               &\textbf{18.69}                                           \\ \hline
\multicolumn{1}{c|}{\multirow{2}{*}{\begin{tabular}[c]{@{}c@{}}OPT-6.7B\\ S=2048\end{tabular}}}  & \multicolumn{1}{c|}{wikitext2}                 & \multicolumn{1}{c?}{10.86}                   & \multicolumn{1}{c|}{11.64}                                                    & \multicolumn{1}{c?}{\textbf{11.57} }                                              & \multicolumn{1}{c|}{14.16}                                                    & \multicolumn{1}{c|}{13.73}                                               & \multicolumn{1}{c|}{14.19}                                               & \textbf{13.72}                                             \\
\multicolumn{1}{c|}{}                                                                            & \multicolumn{1}{c|}{c4}                        & \multicolumn{1}{c?}{12.71}                   & \multicolumn{1}{c|}{13.81}                                                    & \multicolumn{1}{c?}{\textbf{13.77}}                                               & \multicolumn{1}{c|}{16.42}                                                    & \multicolumn{1}{c|}{15.86}                                               & \multicolumn{1}{c|}{16.34}                                               & \textbf{15.85}                                            \\ \hline
\multicolumn{1}{c|}{\multirow{2}{*}{\begin{tabular}[c]{@{}c@{}}OPT-30B\\ S=all\end{tabular}}}    & \multicolumn{1}{c|}{wikitext2}                 & \multicolumn{1}{c?}{9.558}                   & \multicolumn{1}{c|}{9.926}                                                    & \multicolumn{1}{c?}{\textbf{9.824}}                                               & \multicolumn{1}{c|}{10.9}                                                     & \multicolumn{1}{c|}{10.7}                                                & \multicolumn{1}{c|}{10.88}                                               & \textbf{10.7}                                               \\
\multicolumn{1}{c|}{}                                                                            & \multicolumn{1}{c|}{c4}                        & \multicolumn{1}{c?}{11.44}                   & \multicolumn{1}{c|}{12.12}                                                    & \multicolumn{1}{c?}{\textbf{11.98}}                                               & \multicolumn{1}{c|}{13.16}                                                    & \multicolumn{1}{c|}{12.93}                                               & \multicolumn{1}{c|}{13.13}                                               & \textbf{12.93}                                             \\ \hline       \hline    
\multicolumn{1}{c|}{\multirow{2}{*}{\begin{tabular}[c]{@{}c@{}}BLOOM-1.7B\\ S=512\end{tabular}}} & \multicolumn{1}{c|}{wikitext2}                 & \multicolumn{1}{c?}{15.39}                   & \multicolumn{1}{c|}{19.1}                                                     & \multicolumn{1}{c?}{\textbf{18.67} }                                              & \multicolumn{1}{c|}{23.7}                                                     & \multicolumn{1}{c|}{22.91}                                               & \multicolumn{1}{c|}{24.01}                                               & \textbf{22.86}                                              \\
\multicolumn{1}{c|}{}                                                                            & \multicolumn{1}{c|}{c4}                        & \multicolumn{1}{c?}{19.49}                   & \multicolumn{1}{c|}{22.53}                                                    & \multicolumn{1}{c?}{\textbf{22.06}}                                               & \multicolumn{1}{c|}{27.02}                                                    & \multicolumn{1}{c|}{25.95}                                               & \multicolumn{1}{c|}{27.42}                                               & \textbf{25.88}                                               \\ \hline
\multicolumn{1}{c|}{\multirow{2}{*}{\begin{tabular}[c]{@{}c@{}}BLOOM-3B\\ S=2048\end{tabular}}}  & \multicolumn{1}{c|}{wikitext2}                 & \multicolumn{1}{c?}{13.48}                   & \multicolumn{1}{c|}{15.99}                                                    & \multicolumn{1}{c?}{\textbf{15.56} }                                              & \multicolumn{1}{c|}{18.87}                                                    & \multicolumn{1}{c|}{18.6}                                                & \multicolumn{1}{c|}{18.8}                                                & \textbf{18.57 }                                              \\
\multicolumn{1}{c|}{}                                                                            & \multicolumn{1}{c|}{c4}                        & \multicolumn{1}{c?}{17.48}                   & \multicolumn{1}{c|}{19.76}                                                    & \multicolumn{1}{c?}{\textbf{19.3} }                                               & \multicolumn{1}{c|}{22.81}                                                    & \multicolumn{1}{c|}{22.22}                                               & \multicolumn{1}{c|}{22.77}                                               & \textbf{22.2  }                                              \\ \hline
\multicolumn{1}{c|}{\multirow{2}{*}{\begin{tabular}[c]{@{}c@{}}BLOOM-7.1B\\ S=2048\end{tabular}}} & \multicolumn{1}{c|}{wikitext2}                 & \multicolumn{1}{c?}{11.37}                   & \multicolumn{1}{c|}{13}                                                       & \multicolumn{1}{c?}{\textbf{12.86} }                                              & \multicolumn{1}{c|}{14.87}                                                    & \multicolumn{1}{c|}{14.57}                                               & \multicolumn{1}{c|}{14.82}                                               & \textbf{14.57  }                                             \\
\multicolumn{1}{c|}{}                                                                            & \multicolumn{1}{c|}{c4}                        & \multicolumn{1}{c?}{15.2}                    & \multicolumn{1}{c|}{16.71}                                                    & \multicolumn{1}{c?}{\textbf{16.59} }                                              & \multicolumn{1}{c|}{18.79}                                                    & \multicolumn{1}{c|}{18.47}                                               & \multicolumn{1}{c|}{18.74}                                               & \textbf{18.47 }                                              \\ \hline           
\end{tabular}}
\caption{Perplexity comparisons for LLMs with C4 as the calibration dataset. More results are  in Appendix \ref{app:sec:opt} and \ref{app:sec:bloom}. } \label{tab:llama2}
\end{table*}

\subsubsection{Pruning Mask}  

In the algorithm, we  need to select the pruned locations and determine the pruning mask.  

\textbf{Solution $\mathfrak{M}$.} It is too complex to follow Equation (\ref{eq:optimal_loss}) to find out  the pruning mask  with the minimal  pruning  loss. Specifically, it needs to select $k$ weights from all weights for each  combination, leading to  too many combinations. It also needs to compute and sort the losses of all combinations to find out the   minimal loss. Thus, for unstructured pruning, we do not implement Solution $\mathfrak{M}$.    

For semi-structured pruning with N:M sparsity, we implement our Solution $\mathfrak{M}$ based on our optimal loss in Equation~(\ref{eq:optimal_loss}). 
Specifically, in N:M sparsity, we split the weights into groups with M weights in each group, and then select N weights to be pruned in each group. For example, in 2:4 sparsity, there are 2 pruned  weights every 4 weights. Thus, in each group with 4 weights, we   use Equation~(\ref{eq:optimal_loss}) to select 2 weights to be pruned with the minimal loss. In particular,  there are 6 combinations to select 2 elements from 4. We compute the loss with    Equation~(\ref{eq:optimal_loss}) for each combination and find out the minimal loss with its corresponding 2 elements, which are determined to be pruned. By doing this for each group, we can determine the pruning mask for the whole matrix. 

Note that it is still a simplified version of  Equation~(\ref{eq:optimal_loss}),  since each group is computed separately without interactions from other groups. Ideally, Equation~(\ref{eq:optimal_loss}) needs to consider all groups together,  which is unaffordable. For example, if there are $G$ groups with 6 combinations in each group for 2:4 sparsity, there are totally $6^G$ combinations. So we just consider the combinations within each group, without connections between groups.  

\textbf{Solution $\mathfrak{S}$.}  
To reduce the complexity and make the problem tractable,  we can assume that $\bm e_{p\_q_{i}}^T   (2 \bm x \bm x^T )^{-1} \bm e_{p\_q_{i}} $  in Equation (\ref{eq:optimal_loss}) is a diagonal matrix with all zeros for   off-diagonal elements. It means that we ignore the  interactions between multiple pruned locations and each pruned weight does not affect other pruned weights. 
Thus, Equation~(\ref{eq:optimal_loss}) can be transformed to the following,

{\footnotesize
\begin{align} \label{eq:loss_single}
\hat L^* =   \frac{ [\bm w]_{i,j}^2 }{2 [(2 \bm x \bm x^T )^{-1}]_{j,j} }. 
\end{align}}%
We  follow Equation~(\ref{eq:loss_single})   to compute the potential pruning loss  for each single  weight (indexed by ($i$, $j$)).   Then we sort the  losses of all weights and find out the $K$ weights  with smaller losses as the pruned weights. It is similar to the mask searching in SparseGPT~\cite{elias2023sparsegpt}. 

\subsubsection{Optimal Compensation}  \label{sec:optimal_compensation}

With the pruning mask, we need to update  other unpruned weights to compensate the pruning loss.

\textbf{Solution $\mathfrak{M}$.} 
To achieve the best performance, we directly  follow Equation~(\ref{eq:optimal_w}) to compute the modifications of other unpruned weights.  In Equation~(\ref{eq:optimal_w}), we do not need to exactly compute  multiple matrix multiplications such as  $ \bm e_{p\_q_i}^T (2 \bm x \bm x^T )^{-1} $   and  $ \bm e_{p\_q_{i}}^T   (2 \bm x \bm x^T )^{-1} \bm e_{p\_q_{i}}$,  since they just select certain rows or columns in a matrix.  Besides,  the complexity of the inversion  $[\bm e_{p\_q_{i}}^T   (2 \bm x \bm x^T )^{-1} \bm e_{p\_q_{i}} ]^{-1}$   is smaller than $ (2 \bm x \bm x^T )^{-1} $ with a reduced dimension. 

\textbf{Solution $\mathfrak{S}$.} Similar to  the pruning mask, we can reduce the complexity of Equation~(\ref{eq:optimal_w}) by assuming that $\bm e_{p\_q_{i}}^T   (2 \bm x \bm x^T )^{-1} \bm e_{p\_q_{i}} $  in Equation (\ref{eq:optimal_w}) is a diagonal matrix with all zeros for   off-diagonal elements. It ignores the  interactions between  multiple pruned weights, and the solution is  similar to that in SparseGPT \cite{elias2023sparsegpt}. For simplicity, we directly follow SparseGPT for  Solution $\mathfrak{S}$ of optimal compensation.

\subsection{Accurate  Pruning Algorithms} \label{sec:algorithm_detailed}  

We design our post-training pruning algorithms for both unstructured  and semi-structured sparsity. 

\subsubsection{Unstructured Post-Training Pruning}

To align with SparseGPT for a fair comparison, we adopt the block pruning setting. The weight matrix is split into blocks with a number of  $S$  columns (block-size) in each block.  All blocks share the same 
 pruning rate $\alpha$  to keep   overall pruning rate.  

In  Algorithm~\ref{alg:b},  for all blocks,  based on how to solve the  pruning mask and optimal compensation,  we have two combinations, $\mathfrak{S} \mathfrak{S}$  and $\mathfrak{S} \mathfrak{M}$. 
The first $\mathfrak{S} $ (or $\mathfrak{M}$) denotes using Solution $\mathfrak{S} $ (or $\mathfrak{M}$)  for pruning mask, and the second $\mathfrak{S} $ (or $\mathfrak{M}$) represents   Solution $\mathfrak{S} $ (or $\mathfrak{M}$)   for optimal compensation. We do not implement Solution $\mathfrak{M}$  for pruning mask due to its huge complexity. $\mathfrak{S} \mathfrak{S}$  is just  SparseGPT.



For the number of columns $S$ in a block, $S = 1$ leads to too many blocks with high complexity. A typical $S$ value is 128, 512, and 2048. $S = all$  means that all  columns are in the same block. 

\subsubsection{Semi-Structured Post-Training Pruning}

Similarly, in semi-structured pruning, we can use Solution $\mathfrak{S}$ or $\mathfrak{M}$ for pruning mask and optimal compensation, leading to 4 combinations: $\mathfrak{S}\mathfrak{S}$, $\mathfrak{S}\mathfrak{M}$, $\mathfrak{M}\mathfrak{S}$,  and $\mathfrak{M}\mathfrak{M}$.  The first $\mathfrak{S} $ (or $\mathfrak{M}$) denotes using Solution $\mathfrak{S} $ (or $\mathfrak{M}$)  for pruning mask, and the second $\mathfrak{S} $ (or $\mathfrak{M}$) is for optimal compensation.  

\subsection{Comparison with the SRP-based Solution} \label{sec: solution_comp}

As discussed in Sec.~\ref{sec:limitation},  the SRP-based method such as SparseGPT needs to 
freeze (fix) all weights  previous to the current pruned weight, incurring certain performance degradation without further updating the frozen unpruned weights. 
Different from SparseGPT,  our MRP solution  updates all unpruned weights, resulting in better performance.  
Furthermore, ours is not a simple sum of multiple SRP solutions. Instead, it depends on not only the pruned weights, but also the interactions between them, as shown in Equation~(\ref{eq:optimal_w}) and (\ref{eq:optimal_loss}).

\section{Experimental Results}

\begin{table}[t]
\scalebox{0.62}{\begin{threeparttable}
\begin{tabular}{c|c?ccc?ccc}
\hline
\multirow{2}{*}{Model}                                                 & \multirow{2}{*}{Method} & \multicolumn{3}{c?}{Sparsity: 70\%}                                                         & \multicolumn{3}{c}{Sparsity: 80\% }                                                          \\ \cline{3-8} 
                                                                       &                         & \multicolumn{1}{c|}{WT2}            & \multicolumn{1}{c|}{PTB}            & C4             & \multicolumn{1}{c|}{WT2}            & \multicolumn{1}{c|}{PTB}            & C4             \\ \hline
\multirow{3}{*}{\begin{tabular}[c]{@{}c@{}}BLOOM-\\ 7.1B\end{tabular}} & Original                & \multicolumn{1}{c|}{11.37}          & \multicolumn{1}{c|}{20.82}          & 15.2           & \multicolumn{1}{c|}{11.37}          & \multicolumn{1}{c|}{20.82}          & 15.2           \\ \cline{2-8} 
                                                                       & SparseGPT               & \multicolumn{1}{c|}{26.79}          & \multicolumn{1}{c|}{62.24}          & 30.3           & \multicolumn{1}{c|}{150.77}         & \multicolumn{1}{c|}{266.9}          & 121.6          \\
                                                                       & Ours-$\mathfrak{SM}$                   & \multicolumn{1}{c|}{\textbf{22.69}} & \multicolumn{1}{c|}{\textbf{49.35}} & \textbf{25.47} & \multicolumn{1}{c|}{\textbf{93.48}} & \multicolumn{1}{c|}{\textbf{168.2}} & \textbf{70.75} \\ \hline
\multirow{4}{*}{\begin{tabular}[c]{@{}c@{}}LLaMA2-\\ 13B\end{tabular}} & Original                & \multicolumn{1}{c|}{4.884}          & \multicolumn{1}{c|}{50.94}          & 6.727          & \multicolumn{1}{c|}{4.884}          & \multicolumn{1}{c|}{50.94}          & 6.727          \\ \cline{2-8} 
                                                                       & Wanda$^{1}$                   & \multicolumn{1}{c|}{-}               & \multicolumn{1}{c|}{-}               &   -             & \multicolumn{1}{c|}{2e3}               & \multicolumn{1}{c|}{-}               &        -        \\
                                                                       & SparseGPT               & \multicolumn{1}{c|}{26.47}          & \multicolumn{1}{c|}{568}            & 27.81          & \multicolumn{1}{c|}{339.4}          & \multicolumn{1}{c|}{1872}           & 262.9          \\
                                                                       & Ours-$\mathfrak{SM}$                    & \multicolumn{1}{c|}{\textbf{19.05}} & \multicolumn{1}{c|}{\textbf{451.2}} & \textbf{22.12} & \multicolumn{1}{c|}{\textbf{93.43}} & \multicolumn{1}{c|}{\textbf{861.8}} & \textbf{88.36} \\ \hline
\multirow{3}{*}{OPT-66B}                                               & Original                & \multicolumn{1}{c|}{9.339}          & \multicolumn{1}{c|}{13.36}          & 10.99          & \multicolumn{1}{c|}{9.339}          & \multicolumn{1}{c|}{13.36}          & 10.99          \\  \cline{2-8} 
                                                                       & SparseGPT               & \multicolumn{1}{c|}{16.62}          & \multicolumn{1}{c|}{28.14}          & 16.87          & \multicolumn{1}{c|}{1.5e4}          & \multicolumn{1}{c|}{1e4}          & 6e3           \\
                                                                       & Ours-$\mathfrak{SM}$                    & \multicolumn{1}{c|}{\textbf{14.41}} & \multicolumn{1}{c|}{\textbf{23.78}} & \textbf{14.92} & \multicolumn{1}{c|}{\textbf{58.39}} & \multicolumn{1}{c|}{\textbf{147.7}} & \textbf{42.75} \\ \hline
\multirow{4}{*}{\begin{tabular}[c]{@{}c@{}}LLaMA2-\\ 70B\end{tabular}} & Original                & \multicolumn{1}{c|}{3.319}          & \multicolumn{1}{c|}{24.25}          & 5.709          & \multicolumn{1}{c|}{3.319}          & \multicolumn{1}{c|}{24.25}          & 5.709          \\  \cline{2-8} 
                                                                       & Wanda$^{1}$                 & \multicolumn{1}{c|}{-}               & \multicolumn{1}{c|}{-}               &       -         & \multicolumn{1}{c|}{1e2}               & \multicolumn{1}{c|}{-}               &        -        \\
                                                                       & SparseGPT               & \multicolumn{1}{c|}{9.042}          & \multicolumn{1}{c|}{56.36}          & 11.69          & \multicolumn{1}{c|}{30.12}          & \multicolumn{1}{c|}{285.3}          & 33.12          \\
                                                                       & Ours-$\mathfrak{SM}$                    & \multicolumn{1}{c|}{\textbf{8.31}} & \multicolumn{1}{c|}{\textbf{51.69}} & \textbf{11.12} & \multicolumn{1}{c|}{\textbf{26.35}} & \multicolumn{1}{c|}{\textbf{219.5}} & \textbf{28.2} \\ \hline
\end{tabular}
\begin{tablenotes}
\item[1] Wanda is not able to run on a single GPU for  
large LLMs such as LLAMA2-13B and OPT-30B. Its results are from the Wanda paper. 
\end{tablenotes}
\end{threeparttable}}
\caption{Perplexity  comparison with baselines.   WT2 denotes WikiText2. The calibration dataset is C4. More results are shown in Appendix \ref{app:sec:others}. } \label{tab:others}
\end{table}

Our implementation for the proposed method is based on  PyTorch \cite{paszke2019pytorch}  and HuggingFace \cite{wolf2019huggingface}. 
We sequentially prune the linear layers of the blocks in LLMs, which only loads one single block each time with significantly less memory cost  \cite{yao2022zeroquant,frantar2022optimal,elias2023sparsegpt}.   
We conduct   experiments  on one  NVIDIA A100 GPU. Similar to \cite{dettmers2022llm,elias2023sparsegpt}, we  do not incorporate any finetuning.   


\textbf{Models.}  We  test our method for transformer-based LLM families (including  LLaMA2 \cite{touvron2023llama}, OPT \cite{zhang2022opt}, and BLOOM \cite{scao2022bloom}) and Mamba-based LLMs \cite{gu2023mamba}.  For each LLM family, we experiment with  multiple models of different sizes   to demonstrate the general performance.

\textbf{Datasets.}
For the calibration data,  for a fair comparison, we adopt the same setting as SparseGPT  and Wanda \cite{sun2023wanda} to randomly choose 128 segments each with 2048 tokens  from the first shard of the C4 dataset \cite{raffel2020exploring} or the LAMBADA dataset \cite{paperno2016lambada}. 

For  performance evaluation, we test the models on commonly used datasets including   raw-WikiText2  \cite{merity2016pointer},  PTB \cite{marcus1994penn}   and  C4.  
We also test on ZeroShot  datasets including LAMBADA \cite{paperno2016lambada}, HellaSwag \cite{zellers2019hellaswag}, PIQA \cite{tata2003piqa},  ARC-Easy and ARC-Challenge \cite{boratko2018systematic},  and WinoGrade \cite{ai2:winogrande}.

\textbf{Baselines.}  We  compare with SparseGPT, Wanda \cite{sun2023wanda}, and Magnitude \cite{zhu2017prune} methods.  We do not   compare with   AdaPrune  \cite{hubara2021accurate} as it   performs worse than SparseGPT.  

\textbf{Configurations.}  We adopt the \textit{perplexity}  to evaluate the accuracy of sparse models.  
To make a fair comparison,  we adopt the same hyperparameters as SparseGPT,  including  the dampening ratio, the    calibration data number,  token length, and so on.  We also test the accuracy on zero-shot datasets.

\subsection{Results on Transformer-based LLMs}

\begin{table*}[t]
\center
\scalebox{0.69}{
\begin{threeparttable}
\begin{tabular}{c|c|c?c?c|c|c|c|c|c|c}
\hline
\multirow{2}{*}{Model}      & \multirow{2}{*}{Method} & \multirow{2}{*}{Sparsity} & Perplexity $\downarrow$ & \multicolumn{7}{c}{Accuracy $\uparrow$}                                                                                                                                                                        \\ \cline{4-11} 
                            &                         &                           & LAMBADA    & \multicolumn{1}{c|}{LAMBADA} & \multicolumn{1}{c|}{HellaSwag} & \multicolumn{1}{c|}{PIQA}  & \multicolumn{1}{c|}{Arc-E} & \multicolumn{1}{c|}{Arc-C} & \multicolumn{1}{c|}{WinoGrade} &  Average \\ \hline                     
\multirow{4}{*}{Mamba-130M} & Magnitude & 50\%     & 1e20       & 0.19   & 27.21     & 53.92 & 30.18 & 25.94 & 50.59     & 31.338      \\
                            & Sparsegpt & 50\%       & 29.8       & \textbf{35.65}  & \textbf{32.37}     & \textbf{60.88} & \textbf{41.88} & 23.38 & 51.62     & 40.963      \\
                            & Wanda    & 50\%       & 44.99      & 29.38  & 32.1      & 60.28 & 42.47 & 23.89 & 51.14     & 39.877      \\
                            & Ours-$\mathfrak{SM}$      & 50\%      & \textbf{28.97}      & 35.2   & 32.21     & 60.83 & 41.58 & \textbf{24.23} & \textbf{51.93}     & \textbf{40.997}      \\ \hline
\multirow{4}{*}{Mamba-370M} & Magnitude & 50\%       & 9e9        & 2.37   & 29.57     & 55.93 & 29.92 & 24.06 & 50.28     & 32.022      \\
                            & Sparsegpt & 50\%      & 13.39      & 45.8   & 39.71     & \textbf{65.13} & \textbf{48.48} & 26.28 & 53.59     & 46.498      \\
                            & Wanda     & 50\%       & 17.69    & 41.16  & 38.96     & 65.02 & 47.64 & 25    & 53.28     & 45.177      \\
                            & Ours-$\mathfrak{SM}$      &  50\%      & \textbf{12.33 }     & \textbf{47.88}  &\textbf{ 40.21 }    & 64.69 & 47.64 & \textbf{26.54} & \textbf{54.3}      & \textbf{46.877 }     \\ \hline
\multirow{4}{*}{Mamba-790M} & Magnitude & 50\%       & 2e58      & 0.04   & 28.98     & 56.8  & 27.36 & 24.74 & 51.14     & 31.510      \\
                            & Sparsegpt & 50\%       & 8.31       & 54.43  & 47.71     & 68.28 & \textbf{51.94} & 25.17 & 55.8      & 50.555      \\
                            & Wanda     & 50\%     & 11.89    & 48.19  & 46.81     & 68.61 & 52.74 & 26.19 & 53.75     & 49.382      \\
                            & Ours-$\mathfrak{SM}$      & 50\%      & \textbf{7.865}      & \textbf{56.01}  & \textbf{47.96}     & \textbf{68.88} & 51.56 & \textbf{26.28} & \textbf{55.88 }    & \textbf{51.095 }     \\ \hline
\multirow{4}{*}{Mamba-1.4B} & Magnitude & 70\%      & 5e6        & 0.29   & 27.29     & 53.24 & 31.27 & 21.25 & 50.2      & 30.590      \\
                            & Sparsegpt & 70\%      & 31.66      & 34.66  & 34.66     & 61.1  & 40.91 & 22.7  & \textbf{55.01}     & 41.507      \\
                            &Wanda     & 70\%      & 1936    & 4.68   & 28.35     & 56.91 & 35.02 & 21.84 & 51.54     & 33.057      \\
                            & Ours-$\mathfrak{SM}$      & 70\%     & \textbf{19.65}    & \textbf{41.96}  & \textbf{35.74}     & \textbf{61.1}  & \textbf{41.16} & \textbf{22.87} & 54.38     & \textbf{42.868}      \\ \hline
\multirow{3}{*}{Mamba-2.8B} & Sparsegpt & 70\%      & 9.964     & 53.58  & 42.19     & 63.82 & \textbf{46.97} & 24.83 & 56.27     & 47.943      \\
                            & Wanda     & 70\%      & 160.7   & 17.62  & 32.91     & 59.36 & 39.14 & 21.42 & 52.8      & 37.208      \\
                            & Ours-$\mathfrak{SM}$      & 70\%      & \textbf{7.511}     & \textbf{58.82}  & \textbf{43.25}     & \textbf{64.64} & 46.63 & \textbf{25.17} & \textbf{58.25}     & \textbf{49.460}      \\ \hline
\end{tabular}
\end{threeparttable}}
\caption{Results for Mamba models. The calibration dataset is LAMBADA.  Magnitude, Wanda and SparseGPT are not implemented for Mamba models in original papers. We implement and adapt these baselines for Mamba. }\label{tab:mamba}
\end{table*}

The results on transformer-based LLMs  under various block-size settings are presented in Table~\ref{tab:llama2}.  More results for OPT and BLOOM models are shown in Appendix \ref{app:sec:opt} and \ref{app:sec:bloom}. We compare the perplexity for unstructured pruning (50\% sparsity) and semi-structured pruning (2:4 sparsity).   As discussed in Section \ref{sec:algorithm}, 
in unstructured pruning, we only have $\mathfrak{SS}$ and $\mathfrak{SM}$ since the complexity to address pruning mask with Solution $\mathfrak M$ is too high. In semi-structured pruning,  we have 4 combinations, $\mathfrak{SS}$, $\mathfrak{SM}$, $\mathfrak{MS}$, and $\mathfrak{MM}$. 
$\mathfrak{SS}$ corresponds to the original SparseGPT, and  other methods  are based on our proposed optimal solution.


As demonstrated in  Table~\ref{tab:llama2}, our method can  achieve a lower perplexity for various models on different datasets under the same setting. Specifically, for unstructured sparsity, our perplexity is  lower than SparseGPT.  
For 2:4 sparsity, all of our methods  achieve lower perplexity than SparseGPT.  $\mathfrak{MM}$ typically performs the best  since it adopts more advanced techniques following our optimal solution. However, it  has the highest complexity.  We notice that the performance of  $\mathfrak{SM}$ is very close to (or even better than) that of  $\mathfrak{MM}$,  with lower complexity. Thus, we suggest to use $\mathfrak{SM}$ with a limited computation budget. 
In the following, we mainly show the results of $\mathfrak{SM}$.

More results for other models are demonstrated in  Table \ref{app:tab:opt}, \ref{app:tab:bloom} and \ref{app:tab:others} with similar observations.  We also demonstrate the comparison with other baselines under different sparsity in Table \ref{tab:others} and Appendix \ref{app:sec:others}.  Ours can achieve better perplexity.

\subsection{Results for Mamba-based LLMs}

The results for Mamba models \cite{gu2023mamba} are demonstrated in Table~\ref{tab:mamba}. 
Similarly, our method can   achieve a better perplexity than other baselines for various Mamba models under the same setting.

\subsection{Zero-Shot Evaluation}

The zero-shot results for Mamba model are demonstrated in Table~\ref{tab:mamba}.
Our method with a better perplexity  can   achieve a higher average accuracy on zero-shot datasets than other baselines.

In Table~\ref{tab:mamba}, we can observe that with 50\%  sparsity, the accuracy for LAMBADA under Magnitude pruning is very low, such as 2.37\% for Mamba-370M, while it is a bit higher on other datasets such as HellaSwag with 29.57\%. 
This observation highlights that the LAMBADA dataset is quite sensitive to the model sparsity. The reason is that LAMBADA  is a token prediction dataset, while other datasets are based on selection from candidate answers, such as Hellaswag  to choose one from 4 candidates with 25\% accuracy for random guessing.
Thus, with a relatively large sparsity such as 50\%, the magnitude pruned model does not perform well, just similar to or a bit better than random guessing. Then for the token prediction of LAMBADA, it achieves extremely low accuracy such as 2.37\% since random guessing can hardly guess the correct token with too many choices. But its accuracy on HellaSwag can still be 30\%, which is just a bit better than random guessing (25\% with 4 choices). We can see that achieving good performance on the LAMBADA dataset demonstrates the superior performance of our method.

\subsection{Ablation Study}

We ablate different values of the dampening ratios and the number of calibration data. We test our method on the LLaMA2-7B model. 
As shown in Appendix \ref{app:sec:ablation}  and Figure~\ref{app:fig:ablation}, by using  a smaller dampening ratio or more calibration data, our performance can be better. But to make a fair comparison, we   set $\gamma=0.01$ and use 128  samples.

\section{Related Work}

\textbf{Post-training pruning.}  It is challenging to apply the traditional pruning methods for LLMs, since it requires to  retrain or finetune the model on the full dataset for many epochs with massive data and  computation costs \cite{yang2023pruning,zhang2022advancing,li2022pruning,Zhan_2021_ICCV,zheng2024exploring}.  To address this problem,  post-training pruning for LLMs are explored to prune  the model with a small amount of calibration data, requiring much less resources compared with retraining.  The post-training idea is originally proposed  in quantization \cite{nagel2020up,hubara2021accurate,shen2024agile,frantar2022gptq} for  transformers and LLMs, and then successfully applied for pruning \cite{hubara2021accelerated,kwon2022fast,shen2024edgeqat,elias2023sparsegpt,shen2024search}.  Post-training compression   usually  investigates the compression for a single layer of the LLM instead of the whole model  for simplicity.  The memory cost is reduced    significantly as the memory only loads  one block  at  a time \cite{hubara2021accelerated,frantar2022gptq}.


\textbf{Post-training solvers.}  
AdaPrune \cite{hubara2021accelerated} uses weight magnitudes to determine the pruning mask, and then uses an optimizer such as  SGD  to update unpruned weights and improve the performance based on a small amount of calibration data.  Its performance is sub-optimal due to the  limited number of finetuning data.  To further  improve the performance, the optimization-based post-training methods are proposed such as    OBC \cite{frantar2022optimal}  and SparseGPT \cite{elias2023sparsegpt}. Based on the  SRP and its solution \cite{singh2020woodfisher}, OBC introduces a greedy solver to remove  one single weight at a time, and then reconstructs the remaining weights following closed-form solutions in each iteration. SparseGPT   further improves the solution of OBC and applies to the largest available open-source LLM models,  achieving 60\% unstructured sparsity with SOTA performance in perplexity.  Besides   optimization-based methods,  the  heuristic  Wanda \cite{sun2023wanda}   suffers from significant  accuracy loss when the sparsity is large. 


\section{Conclusion}

We first formulate the  MRP  in LLMs to prune multiple weights simultaneously.  Then we derive the optimal solution. Based on the optimal solution, we propose accurate post-training pruning algorithms for unstructured and semi-structured sparsity.  Our comprehensive  experiments demonstrate that our method is more accurate than  SOTA baselines  under the same configurations.


\section*{Limitations}
The complexity of our method may be higher than SparseGPT, as we need to invert the matrix multiple times. As discussed in Sec. \ref{sec:optimal_compensation},  we can avoid certain matrix multiplications with row or column selection, and  the complexity of matrix inversion is reduced due to a smaller matrix dimension.  Our method can still be finished with  one single GPU within a few hours.

\bibliography{anthology,custom}
\bibliographystyle{acl_natbib}

\appendix

\onecolumn

\setcounter{table}{0}
\renewcommand{\thetable}{A\arabic{table}}
\renewcommand*{\theHtable}{\thetable}

\setcounter{figure}{0}
\renewcommand{\thefigure}{A\arabic{figure}}
\renewcommand*{\theHfigure}{\thefigure}

\noindent\textbf{\large Appendix}

\section{Results on OPT Models}
\label{app:sec:opt}

The results for OPT models are demonstrated in Table~\ref{app:tab:opt}. Similarly, our methods can   achieve a better perplexity than SparseGPT for various models on different datasets under the same setting.
For 2:4 sparsity,   $\mathfrak{MM}$ typically can achieve the best performance and  the performance of $\mathfrak{SM}$  is very close to (or even better than) that of $\mathfrak{MM}$  with lower complexity.

\begin{table*}[h]
\scalebox{0.95}{
\begin{tabular}{ccccccccc}
\hline
\multicolumn{1}{c|}{\multirow{2}{*}{\begin{tabular}[c]{@{}c@{}}Model \&\\ Setting\end{tabular}}} & \multicolumn{1}{c|}{\multirow{2}{*}{Datasets}} & \multicolumn{1}{c?}{\multirow{2}{*}{\begin{tabular}[c]{@{}c@{}}Original \\ perplexity\end{tabular}}} & \multicolumn{2}{c?}{Unstructured 50\%}                                                                                                                   & \multicolumn{4}{c}{2:4 sparsity}                                                                                                                                                                                                                                                          \\ \cline{4-9} 
\multicolumn{1}{c|}{}                                                                            & \multicolumn{1}{c|}{}                          & \multicolumn{1}{c?}{}                        & \multicolumn{1}{c|}{\begin{tabular}[c]{@{}c@{}}$\mathfrak{S}\mathfrak{S}$\\ (SparseGPT)\end{tabular}} & \multicolumn{1}{c?}{\begin{tabular}[c]{@{}c@{}} $\mathfrak{S}\mathfrak{M}$\\ (ours)\end{tabular}} & \multicolumn{1}{c|}{\begin{tabular}[c]{@{}c@{}}$\mathfrak{S}\mathfrak{S}$\\ (SparseGPT)\end{tabular}} & \multicolumn{1}{c|}{\begin{tabular}[c]{@{}c@{}}$\mathfrak{S}\mathfrak{M}$\\ (ours)\end{tabular}} & \multicolumn{1}{c|}{\begin{tabular}[c]{@{}c@{}}$\mathfrak{M}\mathfrak{S}$\\ (ours)\end{tabular}} & \begin{tabular}[c]{@{}c@{}}$\mathfrak{M}\mathfrak{M}$\\ (ours)\end{tabular} \\  \hline 
\multicolumn{1}{c|}{\multirow{3}{*}{\begin{tabular}[c]{@{}c@{}}OPT-125M\\ S=128\end{tabular}}}   & \multicolumn{1}{c|}{wikitext2}                 & \multicolumn{1}{c?}{27.65}                   & \multicolumn{1}{c|}{37.02}                                                    & \multicolumn{1}{c?}{\textbf{35.75} }                                              & \multicolumn{1}{c|}{58.78}                                                    & \multicolumn{1}{c|}{52.4}                                                & \multicolumn{1}{c|}{59.13}                                               & \textbf{52.31}                                               \\
\multicolumn{1}{c|}{}                                                                            & \multicolumn{1}{c|}{ptb}                       & \multicolumn{1}{c?}{38.99}                   & \multicolumn{1}{c|}{55.4}                                                     & \multicolumn{1}{c?}{\textbf{55.37} }                                              & \multicolumn{1}{c|}{92.42}                                                    & \multicolumn{1}{c|}{\textbf{83.22}  }                                             & \multicolumn{1}{c|}{91.68}                                               & 83.23                                               \\
\multicolumn{1}{c|}{}                                                                            & \multicolumn{1}{c|}{c4}                        & \multicolumn{1}{c?}{26.56}                   & \multicolumn{1}{c|}{33.49}                                                    & \multicolumn{1}{c?}{\textbf{32.72}}                                               & \multicolumn{1}{c|}{51.49}                                                    & \multicolumn{1}{c|}{45.76}                                               & \multicolumn{1}{c|}{49.64}                                               & \textbf{45.59  }                                             \\ \hline
\multicolumn{1}{c|}{\multirow{3}{*}{\begin{tabular}[c]{@{}c@{}}OPT-350M\\ S=128\end{tabular}}}   & \multicolumn{1}{c|}{wikitext2}                 & \multicolumn{1}{c?}{22}                      & \multicolumn{1}{c|}{31.21}                                                    & \multicolumn{1}{c?}{\textbf{30.12} }                                              & \multicolumn{1}{c|}{50.57}                                                    & \multicolumn{1}{c|}{50.06}                                               & \multicolumn{1}{c|}{51.94}                                               & \textbf{48.46    }                                           \\
\multicolumn{1}{c|}{}                                                                            & \multicolumn{1}{c|}{ptb}                       & \multicolumn{1}{c?}{31.07}                   & \multicolumn{1}{c|}{43.44}                                                    & \multicolumn{1}{c?}{\textbf{42.71}  }                                             & \multicolumn{1}{c|}{72.45}                                                    & \multicolumn{1}{c|}{\textbf{70.8}}                                                & \multicolumn{1}{c|}{72.23}                                               & 70.96                                               \\
\multicolumn{1}{c|}{}                                                                            & \multicolumn{1}{c|}{c4}                        & \multicolumn{1}{c?}{22.59}                   & \multicolumn{1}{c|}{29.17}                                                    & \multicolumn{1}{c?}{\textbf{28.36} }                                              & \multicolumn{1}{c|}{46.48}                                                    & \multicolumn{1}{c|}{42.79}                                               & \multicolumn{1}{c|}{46.04}                                               &\textbf{ 42.16  }                                             \\ \hline
\multicolumn{1}{c|}{\multirow{3}{*}{\begin{tabular}[c]{@{}c@{}}OPT-2.7B\\ S=512\end{tabular}}}   & \multicolumn{1}{c|}{wikitext2}                 & \multicolumn{1}{c?}{12.47}                   & \multicolumn{1}{c|}{13.43}                                                    & \multicolumn{1}{c?}{\textbf{13.29}  }                                             & \multicolumn{1}{c|}{17.13}                                                    & \multicolumn{1}{c|}{16.74}                                               & \multicolumn{1}{c|}{16.89}                                               & \textbf{16.68  }                                             \\
\multicolumn{1}{c|}{}                                                                            & \multicolumn{1}{c|}{ptb}                       & \multicolumn{1}{c?}{17.97}                   & \multicolumn{1}{c|}{20.45}                                                    & \multicolumn{1}{c?}{\textbf{20.28}  }                                             & \multicolumn{1}{c|}{26.97}                                                    & \multicolumn{1}{c|}{25.99}                                               & \multicolumn{1}{c|}{26.63}                                               & \textbf{25.91  }                                             \\
\multicolumn{1}{c|}{}                                                                            & \multicolumn{1}{c|}{c4}                        & \multicolumn{1}{c?}{14.34}                   & \multicolumn{1}{c|}{15.8}                                                     & \multicolumn{1}{c?}{\textbf{15.66} }                                              & \multicolumn{1}{c|}{19.34}                                                    & \multicolumn{1}{c|}{18.7}                                                & \multicolumn{1}{c|}{19.07}                                               &\textbf{ 18.69    }                                           \\ \hline
\multicolumn{1}{c|}{\multirow{3}{*}{\begin{tabular}[c]{@{}c@{}}OPT-6.7B\\ S=2048\end{tabular}}}  & \multicolumn{1}{c|}{wikitext2}                 & \multicolumn{1}{c?}{10.86}                   & \multicolumn{1}{c|}{11.64}                                                    & \multicolumn{1}{c?}{\textbf{11.57} }                                              & \multicolumn{1}{c|}{14.16}                                                    & \multicolumn{1}{c|}{13.73}                                               & \multicolumn{1}{c|}{14.19}                                               & \textbf{13.72  }                                             \\
\multicolumn{1}{c|}{}                                                                            & \multicolumn{1}{c|}{ptb}                       & \multicolumn{1}{c?}{15.77}                   & \multicolumn{1}{c|}{17.45}                                                    & \multicolumn{1}{c?}{\textbf{17.33} }                                              & \multicolumn{1}{c|}{21.53}                                                    & \multicolumn{1}{c|}{\textbf{20.38}}                                               & \multicolumn{1}{c|}{21.11}                                               & 20.4                                                \\
\multicolumn{1}{c|}{}                                                                            & \multicolumn{1}{c|}{c4}                        & \multicolumn{1}{c?}{12.71}                   & \multicolumn{1}{c|}{13.81}                                                    & \multicolumn{1}{c?}{\textbf{13.77}}                                               & \multicolumn{1}{c|}{16.42}                                                    & \multicolumn{1}{c|}{15.86}                                               & \multicolumn{1}{c|}{16.34}                                               & \textbf{15.85   }                                            \\ \hline
\multicolumn{1}{c|}{\multirow{3}{*}{\begin{tabular}[c]{@{}c@{}}OPT-30B\\ S=all\end{tabular}}}    & \multicolumn{1}{c|}{wikitext2}                 & \multicolumn{1}{c?}{9.558}                   & \multicolumn{1}{c|}{9.926}                                                    & \multicolumn{1}{c?}{\textbf{9.824}}                                               & \multicolumn{1}{c|}{10.9}                                                     & \multicolumn{1}{c|}{10.7}                                                & \multicolumn{1}{c|}{10.88}                                               & \textbf{10.7 }                                               \\
\multicolumn{1}{c|}{}                                                                            & \multicolumn{1}{c|}{ptb}                       & \multicolumn{1}{c?}{14.04}                   & \multicolumn{1}{c|}{15.3}                                                     & \multicolumn{1}{c?}{\textbf{15.05}}                                               & \multicolumn{1}{c|}{16.58}                                                    & \multicolumn{1}{c|}{\textbf{16.19} }                                              & \multicolumn{1}{c|}{16.53}                                               & 16.2                                                \\
\multicolumn{1}{c|}{}                                                                            & \multicolumn{1}{c|}{c4}                        & \multicolumn{1}{c?}{11.44}                   & \multicolumn{1}{c|}{12.12}                                                    & \multicolumn{1}{c?}{\textbf{11.98}}                                               & \multicolumn{1}{c|}{13.16}                                                    & \multicolumn{1}{c|}{12.93}                                               & \multicolumn{1}{c|}{13.13}                                               & \textbf{12.93  }                                             \\ \hline                        
\end{tabular}}
\caption{Perplexity comparisons for OPT models under various block-size settings.}  \label{app:tab:opt}
\end{table*}

\section{Results on BLOOM Models}
\label{app:sec:bloom}
The results for BLOOM models are demonstrated in Table~\ref{app:tab:bloom}. Similarly, our methods can  achieve a better perplexity than SparseGPT for various models on different datasets under the same setting.
For 2:4 sparsity,    the performance of $\mathfrak{SM}$ is very close to  (or even better than) that of $\mathfrak{MM}$   with lower complexity.

\begin{table*}[h]
\scalebox{0.92}{
\begin{tabular}{ccccccccc}
\hline
\multicolumn{1}{c|}{\multirow{2}{*}{\begin{tabular}[c]{@{}c@{}}Model \&\\ Setting\end{tabular}}} & \multicolumn{1}{c|}{\multirow{2}{*}{Datasets}} & \multicolumn{1}{c?}{\multirow{2}{*}{\begin{tabular}[c]{@{}c@{}}Original \\ perplexity\end{tabular}}} & \multicolumn{2}{c?}{Unstructured 50\%}                                                                                                                   & \multicolumn{4}{c}{2:4 sparsity}                                                                                                                                                                                                                                                          \\ \cline{4-9} 
\multicolumn{1}{c|}{}                                                                            & \multicolumn{1}{c|}{}                          & \multicolumn{1}{c?}{}                        & \multicolumn{1}{c|}{\begin{tabular}[c]{@{}c@{}}$\mathfrak{S}\mathfrak{S}$\\ (SparseGPT)\end{tabular}} & \multicolumn{1}{c?}{\begin{tabular}[c]{@{}c@{}} $\mathfrak{S}\mathfrak{M}$\\ (ours)\end{tabular}} & \multicolumn{1}{c|}{\begin{tabular}[c]{@{}c@{}}$\mathfrak{S}\mathfrak{S}$\\ (SparseGPT)\end{tabular}} & \multicolumn{1}{c|}{\begin{tabular}[c]{@{}c@{}}$\mathfrak{S}\mathfrak{M}$\\ (ours)\end{tabular}} & \multicolumn{1}{c|}{\begin{tabular}[c]{@{}c@{}}$\mathfrak{M}\mathfrak{S}$\\ (ours)\end{tabular}} & \begin{tabular}[c]{@{}c@{}}$\mathfrak{M}\mathfrak{M}$\\ (ours)\end{tabular} \\  \hline 
\multicolumn{1}{c|}{\multirow{3}{*}{\begin{tabular}[c]{@{}c@{}}BLOOM-560M\\ S=128\end{tabular}}} & \multicolumn{1}{c|}{wikitext2}                 & \multicolumn{1}{c?}{22.41}                   & \multicolumn{1}{c|}{29.12}                                                    & \multicolumn{1}{c?}{\textbf{28.77} }                                              & \multicolumn{1}{c|}{37.58}                                                    & \multicolumn{1}{c|}{36.28}                                               & \multicolumn{1}{c|}{38.78}                                               & \textbf{36.02  }                                             \\
\multicolumn{1}{c|}{}                                                                            & \multicolumn{1}{c|}{ptb}                       & \multicolumn{1}{c?}{43.66}                   & \multicolumn{1}{c|}{60.52}                                                    & \multicolumn{1}{c?}{\textbf{60.36} }                                              & \multicolumn{1}{c|}{77.53}                                                    & \multicolumn{1}{c|}{73.93}                                               & \multicolumn{1}{c|}{79.4}                                                & \textbf{73.01 }                                              \\
\multicolumn{1}{c|}{}                                                                            & \multicolumn{1}{c|}{c4}                        & \multicolumn{1}{c?}{26.59}                   & \multicolumn{1}{c|}{32.83}                                                    & \multicolumn{1}{c?}{\textbf{32.12} }                                              & \multicolumn{1}{c|}{40.72}                                                    & \multicolumn{1}{c|}{39.6}                                                & \multicolumn{1}{c|}{41.98}                                               & \textbf{39.46 }                                              \\ \hline
\multicolumn{1}{c|}{\multirow{3}{*}{\begin{tabular}[c]{@{}c@{}}BLOOM-1.7B\\ S=512\end{tabular}}} & \multicolumn{1}{c|}{wikitext2}                 & \multicolumn{1}{c?}{15.39}                   & \multicolumn{1}{c|}{19.1}                                                     & \multicolumn{1}{c?}{\textbf{18.67} }                                              & \multicolumn{1}{c|}{23.7}                                                     & \multicolumn{1}{c|}{22.91}                                               & \multicolumn{1}{c|}{24.01}                                               & \textbf{22.86 }                                              \\
\multicolumn{1}{c|}{}                                                                            & \multicolumn{1}{c|}{ptb}                       & \multicolumn{1}{c?}{29.99}                   & \multicolumn{1}{c|}{39.23}                                                    & \multicolumn{1}{c?}{\textbf{39.21} }                                              & \multicolumn{1}{c|}{48.65}                                                    & \multicolumn{1}{c|}{45.79}                                               & \multicolumn{1}{c|}{48.2}                                                & \textbf{45.41   }                                            \\
\multicolumn{1}{c|}{}                                                                            & \multicolumn{1}{c|}{c4}                        & \multicolumn{1}{c?}{19.49}                   & \multicolumn{1}{c|}{22.53}                                                    & \multicolumn{1}{c?}{\textbf{22.06}}                                               & \multicolumn{1}{c|}{27.02}                                                    & \multicolumn{1}{c|}{25.95}                                               & \multicolumn{1}{c|}{27.42}                                               & \textbf{ 25.88}                                               \\ \hline
\multicolumn{1}{c|}{\multirow{3}{*}{\begin{tabular}[c]{@{}c@{}}BLOOM-3B\\ S=2048\end{tabular}}}  & \multicolumn{1}{c|}{wikitext2}                 & \multicolumn{1}{c?}{13.48}                   & \multicolumn{1}{c|}{15.99}                                                    & \multicolumn{1}{c?}{\textbf{15.56} }                                              & \multicolumn{1}{c|}{18.87}                                                    & \multicolumn{1}{c|}{18.6}                                                & \multicolumn{1}{c|}{18.8}                                                & \textbf{18.57 }                                              \\
\multicolumn{1}{c|}{}                                                                            & \multicolumn{1}{c|}{ptb}                       & \multicolumn{1}{c?}{25.34}                   & \multicolumn{1}{c|}{30.47}                                                    & \multicolumn{1}{c?}{\textbf{29.56} }                                              & \multicolumn{1}{c|}{38.44}                                                    & \multicolumn{1}{c|}{\textbf{37.34}  }                                             & \multicolumn{1}{c|}{38.91}                                               & 37.48                                               \\
\multicolumn{1}{c|}{}                                                                            & \multicolumn{1}{c|}{c4}                        & \multicolumn{1}{c?}{17.48}                   & \multicolumn{1}{c|}{19.76}                                                    & \multicolumn{1}{c?}{\textbf{19.3} }                                               & \multicolumn{1}{c|}{22.81}                                                    & \multicolumn{1}{c|}{22.22}                                               & \multicolumn{1}{c|}{22.77}                                               & \textbf{22.2  }                                              \\ \hline
\multicolumn{1}{c|}{\multirow{3}{*}{\begin{tabular}[c]{@{}c@{}}BLOOM7.1B\\ S=2048\end{tabular}}} & \multicolumn{1}{c|}{wikitext2}                 & \multicolumn{1}{c?}{11.37}                   & \multicolumn{1}{c|}{13}                                                       & \multicolumn{1}{c?}{\textbf{12.86} }                                              & \multicolumn{1}{c|}{14.87}                                                    & \multicolumn{1}{c|}{14.57}                                               & \multicolumn{1}{c|}{14.82}                                               & \textbf{14.57  }                                             \\
\multicolumn{1}{c|}{}                                                                            & \multicolumn{1}{c|}{ptb}                       & \multicolumn{1}{c?}{20.82}                   & \multicolumn{1}{c|}{24.26}                                                    & \multicolumn{1}{c?}{\textbf{23.97} }                                              & \multicolumn{1}{c|}{28.28}                                                    & \multicolumn{1}{c|}{27.86}                                               & \multicolumn{1}{c|}{28.5}                                                &\textbf{ 27.8 }                                               \\
\multicolumn{1}{c|}{}                                                                            & \multicolumn{1}{c|}{c4}                        & \multicolumn{1}{c?}{15.2}                    & \multicolumn{1}{c|}{16.71}                                                    & \multicolumn{1}{c?}{\textbf{16.59} }                                              & \multicolumn{1}{c|}{18.79}                                                    & \multicolumn{1}{c|}{18.47}                                               & \multicolumn{1}{c|}{18.74}                                               & \textbf{18.47 }                                              \\ \hline                           
\end{tabular}}
\caption{Perplexity comparisons for BLOOM models under various block-size settings.}   \label{app:tab:bloom}
\end{table*}

\section{Results of Other Sparsity and Baselines}
\label{app:sec:others}

  We  demonstrate the comparison with other baselines under different sparsity in Table \ref{app:tab:others}.  Our method can achieve better perplexity.

\begin{table*}[h]
\caption{Perplexity  comparison of our method and baselines.   WT2 denotes WikiText2.  Wanda is not able to run on a single GPU for large LLMs such as LLAMA2-13B and OPT-30B. Its results are from the Wanda paper. }
\label{app:tab:others}
\begin{tabular}{c|c|c?ccc?ccc}
\hline
\multirow{2}{*}{Model}      & \multirow{2}{*}{Method} & \multirow{2}{*}{block-size} & \multicolumn{3}{c?}{Sparsity: 0.7}                                                         & \multicolumn{3}{c}{Sparsity: 0.8}                                                           \\ \cline{4-9} 
                            &                         &                             & \multicolumn{1}{c|}{WT2}            & \multicolumn{1}{c|}{PTB}            & C4             & \multicolumn{1}{c|}{WT2}            & \multicolumn{1}{c|}{PTB}             & C4             \\ \hline
\multirow{3}{*}{OPT-6.7B}   & Original                & -                           & \multicolumn{1}{c|}{10.86}          & \multicolumn{1}{c|}{15.77}          & 12.71          & \multicolumn{1}{c|}{10.86}          & \multicolumn{1}{c|}{15.77}           & 12.71          \\
                            & SparseGPT               & 512                         & \multicolumn{1}{c|}{20.7}           & \multicolumn{1}{c|}{31.3}           & 21.68          & \multicolumn{1}{c|}{84.43}          & \multicolumn{1}{c|}{103.6}           & 71.8           \\
                            & Ours-$\mathfrak{SM}$                    & 512                         & \multicolumn{1}{c|}{\textbf{19.84}} & \multicolumn{1}{c|}{\textbf{31.06}} & \textbf{21.18} & \multicolumn{1}{c|}{\textbf{80.52}} & \multicolumn{1}{c|}{\textbf{101.34}} & \textbf{69.2}  \\ \hline
\multirow{4}{*}{LLaMA2-7B}  & Original                & -                           & \multicolumn{1}{c|}{5.472}          & \multicolumn{1}{c|}{37.91}          & 7.263          & \multicolumn{1}{c|}{5.472}          & \multicolumn{1}{c|}{37.91}           & 7.263          \\
                            & Wanda                   & -                           & \multicolumn{1}{c|}{-}              & \multicolumn{1}{c|}{-}              & -              & \multicolumn{1}{c|}{1e5}            & \multicolumn{1}{c|}{-}               & -              \\
                            & SparseGPT               & 512                         & \multicolumn{1}{c|}{26.25}          & \multicolumn{1}{c|}{2203}           & 28.49          & \multicolumn{1}{c|}{104}            & \multicolumn{1}{c|}{4358}            & 104.6          \\
                            & Ours-$\mathfrak{SM}$                    & 512                         & \multicolumn{1}{c|}{\textbf{24.53}} & \multicolumn{1}{c|}{\textbf{1812}}  & \textbf{26.75} & \multicolumn{1}{c|}{\textbf{91.92}} & \multicolumn{1}{c|}{\textbf{3422}}   & \textbf{90.12} \\ \hline
\multirow{3}{*}{BLOOM-7.1B} & Original                & -                           & \multicolumn{1}{c|}{11.37}          & \multicolumn{1}{c|}{20.82}          & 15.2           & \multicolumn{1}{c|}{11.37}          & \multicolumn{1}{c|}{20.82}           & 15.2           \\
                            & SparseGPT               & 2048                        & \multicolumn{1}{c|}{26.79}          & \multicolumn{1}{c|}{62.24}          & 30.3           & \multicolumn{1}{c|}{150.77}         & \multicolumn{1}{c|}{266.9}           & 121.6          \\
                            & Ours-$\mathfrak{SM}$                    & 2048                        & \multicolumn{1}{c|}{\textbf{22.69}} & \multicolumn{1}{c|}{\textbf{49.35}} & \textbf{25.47} & \multicolumn{1}{c|}{\textbf{93.48}} & \multicolumn{1}{c|}{\textbf{168.2}}  & \textbf{70.75} \\ \hline
\multirow{4}{*}{LLaMA2-13B} & Original                & -                           & \multicolumn{1}{c|}{4.884}          & \multicolumn{1}{c|}{50.94}          & 6.727          & \multicolumn{1}{c|}{4.884}          & \multicolumn{1}{c|}{50.94}           & 6.727          \\
                            & Wanda                   & -                           & \multicolumn{1}{c|}{-}              & \multicolumn{1}{c|}{-}              & -              & \multicolumn{1}{c|}{2e3}            & \multicolumn{1}{c|}{-}               & -              \\
                            & SparseGPT               & all                         & \multicolumn{1}{c|}{26.47}          & \multicolumn{1}{c|}{568}            & 27.81          & \multicolumn{1}{c|}{339.4}          & \multicolumn{1}{c|}{1872}            & 262.9          \\
                            & Ours-$\mathfrak{SM}$                    & all                         & \multicolumn{1}{c|}{\textbf{19.05}} & \multicolumn{1}{c|}{\textbf{451.2}} & \textbf{22.12} & \multicolumn{1}{c|}{\textbf{93.43}} & \multicolumn{1}{c|}{\textbf{861.8}}  & \textbf{88.36} \\ \hline
\multirow{3}{*}{OPT-30B}    & Original                & -                           & \multicolumn{1}{c|}{9.558}          & \multicolumn{1}{c|}{14.04}          & 11.44          & \multicolumn{1}{c|}{9.558}          & \multicolumn{1}{c|}{14.04}           & 11.44          \\
                            & SparseGPT               & all                         & \multicolumn{1}{c|}{15.42}          & \multicolumn{1}{c|}{25.63}          & 17.09          & \multicolumn{1}{c|}{604.4}          & \multicolumn{1}{c|}{303.7}           & 349            \\
                            & Ours-$\mathfrak{SM}$                    & all                         & \multicolumn{1}{c|}{\textbf{13.54}} & \multicolumn{1}{c|}{\textbf{22.42}} & \textbf{15.56} & \multicolumn{1}{c|}{\textbf{50.61}} & \multicolumn{1}{c|}{\textbf{71.98}}  & \textbf{39.15} \\ \hline
\multirow{3}{*}{OPT-66B}    & Original                & -                           & \multicolumn{1}{c|}{9.339}          & \multicolumn{1}{c|}{13.36}          & 10.99          & \multicolumn{1}{c|}{9.339}          & \multicolumn{1}{c|}{13.36}           & 10.99          \\
                             & SparseGPT          & all     & \multicolumn{1}{c|}{16.62}          & \multicolumn{1}{c|}{28.14}          & 16.87          & \multicolumn{1}{c|}{1.5e4}          & \multicolumn{1}{c|}{1e4}          & 6e3           \\
                                                                       & Ours-$\mathfrak{SM}$    & all                & \multicolumn{1}{c|}{\textbf{14.41}} & \multicolumn{1}{c|}{\textbf{23.78}} & \textbf{14.92} & \multicolumn{1}{c|}{\textbf{58.39}} & \multicolumn{1}{c|}{\textbf{147.7}} & \textbf{42.75} \\ \hline
\multirow{4}{*}{LLaMA2-70B} & Original                & -                           & \multicolumn{1}{c|}{3.319}          & \multicolumn{1}{c|}{24.25}          & 5.709          & \multicolumn{1}{c|}{3.319}          & \multicolumn{1}{c|}{24.25}           & 5.709          \\
                            & Wanda                   & -                           & \multicolumn{1}{c|}{-}              & \multicolumn{1}{c|}{-}              & -              & \multicolumn{1}{c|}{1e2}            & \multicolumn{1}{c|}{-}               & -              \\
& SparseGPT               &  all & \multicolumn{1}{c|}{9.042}          & \multicolumn{1}{c|}{56.36}          & 11.69          & \multicolumn{1}{c|}{30.12}          & \multicolumn{1}{c|}{285.3}          & 33.12          \\
                                                                       & Ours-$\mathfrak{SM}$                     & all & \multicolumn{1}{c|}{\textbf{8.31}} & \multicolumn{1}{c|}{\textbf{51.69}} & \textbf{11.12} & \multicolumn{1}{c|}{\textbf{26.35}} & \multicolumn{1}{c|}{\textbf{219.5}} & \textbf{28.2} \\ \hline
\end{tabular}
\end{table*}

\section{Ablation Study}
\label{app:sec:ablation}

\begin{figure}[h]    
 \centering
 \scalebox{1.1}{
\begin{tabular}{p{2.1in}p{2.1in}}
\includegraphics[width=2.1in]{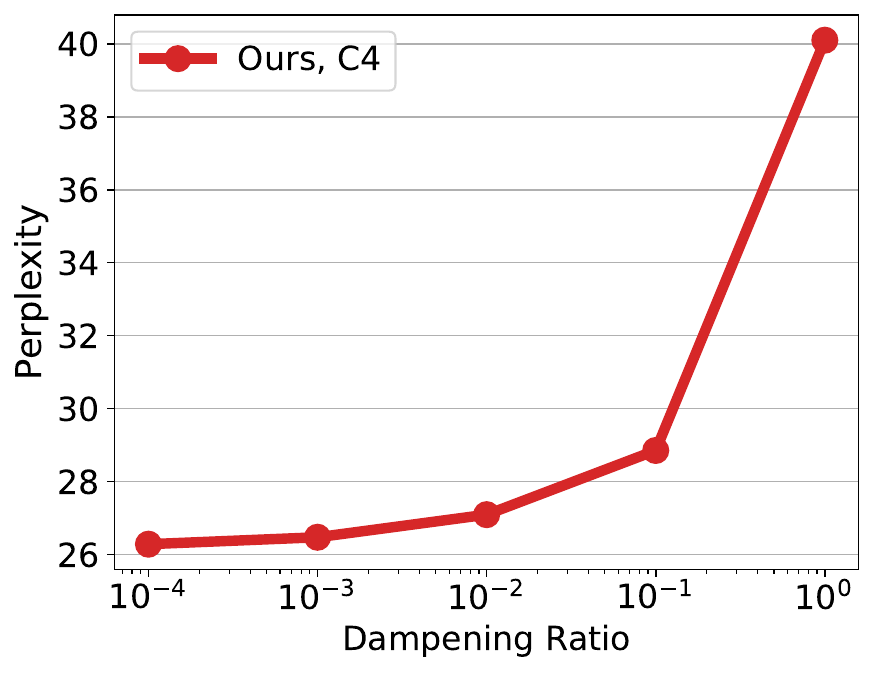}&
\includegraphics[width=2.1in]{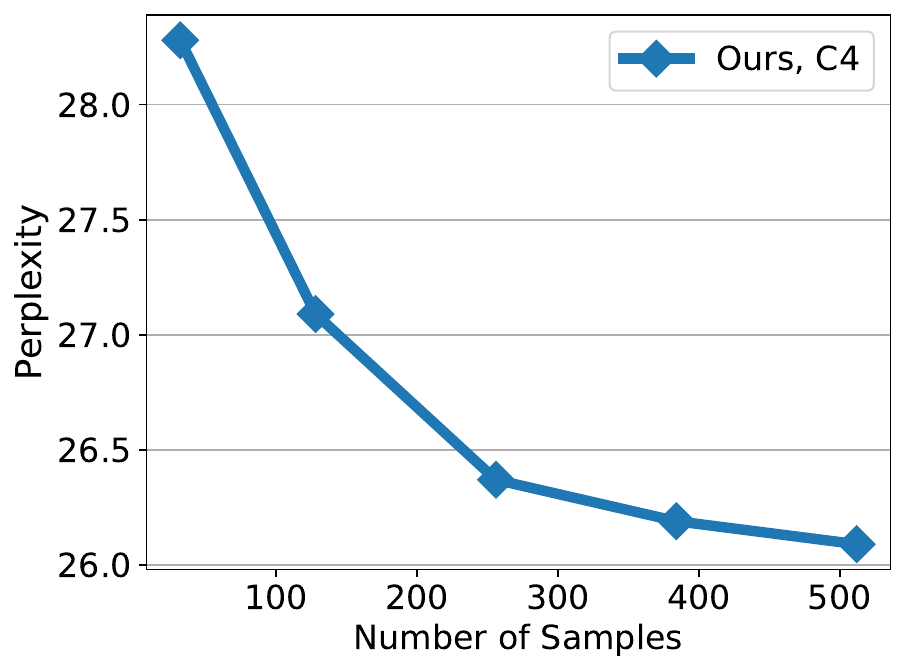}   
\end{tabular}}
\caption{\footnotesize{Ablation study for the dampening ratio and the number of samples. 
}} \label{app:fig:ablation}
\end{figure}

We ablate different values of the dampening ratios and the number of calibration data. We test our $\mathfrak{SM}$ method on the LLAMA2-7B model. 
As shown in  Figure~\ref{app:fig:ablation}, by using  a smaller dampening ratio or more calibration data, our performance can be better. But to make a fair comparison, we   set $\gamma=0.01$ and use 128  samples. 

\end{document}